\documentclass{article}
\usepackage{ijcai26}

\usepackage{times}
\usepackage{soul}
\usepackage{url}
\usepackage[hidelinks]{hyperref}
\usepackage[utf8]{inputenc}
\usepackage[small]{caption}
\usepackage{graphicx}
\usepackage{amsmath}
\usepackage{amsthm}
\usepackage{booktabs}
\usepackage{algorithm}
\usepackage{algorithmic}
\usepackage[switch]{lineno}
\usepackage{amssymb}
\usepackage{multirow}
\usepackage{pifont}
\usepackage[table]{xcolor}
\usepackage{mathtools}
\usepackage{subfigure} 
\usepackage{subcaption}
\usepackage{fontawesome5}
\definecolor{deeppurple}{HTML}{5B2C83}

\title{Bi-CoG: Bi-Consistency-Guided Self-Training for Vision-Language Models}

\author{
Rui Zhu$^{1*}$
\and
Song-Lin Lv$^{1,2*}$\and
Zi-Kang Wang$^{1,2}$\And
Lan-Zhe Guo$^{1,2\dagger}$\\
\affiliations
$^1$School of Intelligence Science and Technology, Nanjing University, China\\
$^2$National Key Laboratory for Novel Software Technology, Nanjing University, China\\
\emails
zhurui@smail.nju.edu.cn, \{lvsl,wangzk,guolz\}@lamda.nju.edu.cn
}

\begin{document}

\maketitle

\renewcommand*{\thefootnote}{\fnsymbol{footnote}}
\footnotetext[1]{Equal contribution.}
\footnotetext[2]{Corresponding author.}
\renewcommand*{\thefootnote}{\arabic{footnote}}

\begin{abstract}
    Exploiting unlabeled data through semi-supervised learning (SSL) or leveraging pre-trained models via fine-tuning are two prevailing paradigms for addressing label-scarce scenarios. Recently, growing attention has been given to combining fine-tuning of pre-trained vision-language models (VLMs) with SSL, forming the emerging paradigm of semi-supervised fine-tuning. However, existing methods often suffer from model bias and hyperparameter sensitivity, due to reliance on prediction consistency or pre-defined thresholds. To address these limitations, we propose a simple yet effective plug-and-play method named \underline{\textbf{Bi-Co}}nsistency-\underline{\textbf{G}}uided Self-Training (Bi-CoG), which assigns accurate and low-bias pseudo-labels, by simultaneously exploiting inter- and intra-model consistency, along with an error-aware dynamic filtering strategy. Both theoretical analysis and extensive experiments over 14 datasets demonstrate the effectiveness of Bi-CoG, which consistently and significantly improves the performance of existing methods. Codes are available at~\href{https://github.com/rayzhu-cmu/Bi-CoG}{\faGithub~\textcolor{deeppurple}{Bi-CoG}}.
\end{abstract}
\section{Introduction}

Addressing downstream tasks with limited labeled data remains a critical challenge in real-world applications~\cite{yang2022sslsurvey,shao2026chinatravel,yang2025neurosymbolic}.
Two prominent paradigms have emerged to tackle this issue: (1) adapting pre-trained models, such as vision-language models (VLMs) \cite{radford2021clip,jia2021align}, for downstream tasks; and (2) applying semi-supervised learning (SSL) methods that leverage large amounts of inexpensive unlabeled data. Recent work \cite{unlabel_or_pretrain} has further suggested combining them, where the strong zero-shot capabilities of pre-trained VLMs are used to generate reliable pseudo-labels that guide semi-supervised fine-tuning, thereby fully exploiting the strengths of both paradigms.
\begin{figure}[t]
\centering
\includegraphics[width=0.95\columnwidth]{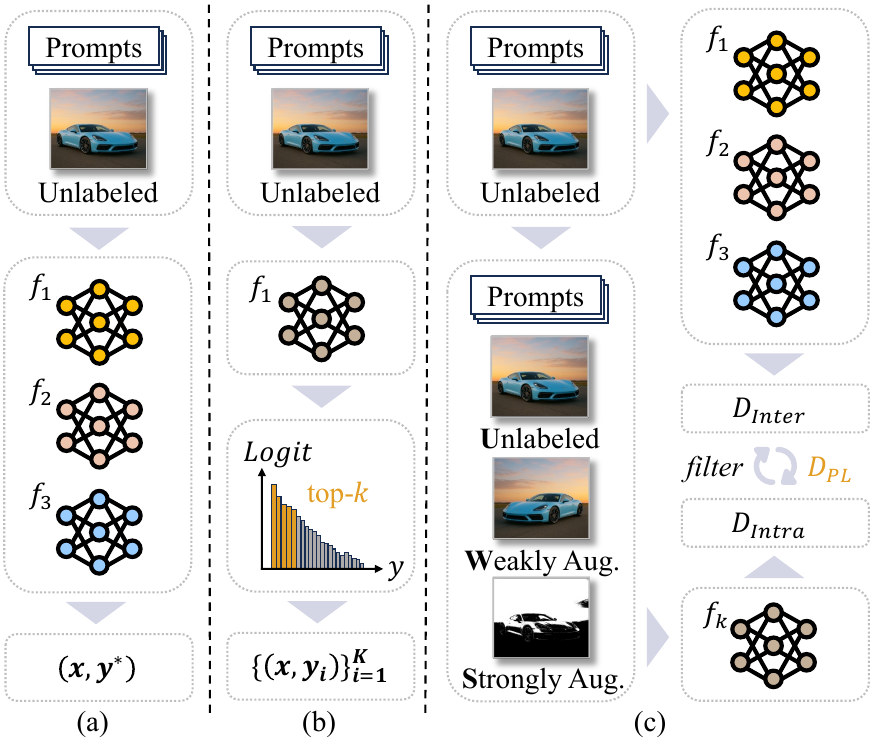} 
\caption{Comparison of existing pseudo-labeling paradigms and Bi-CoG. (a) Majority voting from multiple VLMs; (b) Top-$k$ selection based on predicted logits; (c) Bi-CoG dynamically generates pseudo-labels based on bi-consistency and model error.}
\label{intro}
\end{figure}

Recent efforts have explored integrating these two paradigms by leveraging pre-trained models to extract domain knowledge from unlabeled data for downstream task adaptation. A common approach is model ensembling (Figure \ref{intro}.a), such as V-PET \cite{gupte2024vpet}, which aggregates multi-view one-hot predictions from multiple vision models. As shown in Figure~\ref{intro}.b, another line of work selects top-$k$ predictions using class-wise or confidence-based thresholds. For instance, GRIP~\cite{menghini2023grip} and UPL~\cite{huang2022upl} enforce class balance by selecting equal samples per class; CPL~\cite{zhang2024cpl} refines labels via instance-level confidence matrices.

However, these methods struggle to balance the trade-off between pseudo-label accuracy and bias. Ensemble-based approaches can produce highly accurate pseudo-labels, but they often inadvertently amplify biases inherited from pre-training, leading to uneven performance across subgroups—a phenomenon commonly referred to as the \textit{``Matthew effect''} \cite{rich_get_richer,debiased_self_training,erasing-the-bias,guo2020safedeep}. In contrast, soft-labeling techniques and class-wise pseudo-label allocation strategies can alleviate overconfidence to some extent, but they tend to introduce additional label noise and are highly sensitive to manually tuned hyperparameters. To the best of our knowledge, adaptively generating reliable and unbiased pseudo-labels in response to model performance remains a critical yet underexplored challenge in semi-supervised fine-tuning.

To address this challenge, we propose a novel method named \underline{\textbf{Bi-Co}}nsistency-\underline{\textbf{G}}uided Self-Training (Bi-CoG). As illustrated in Figure~\ref{intro}.c, Bi-CoG leverages multiple VLMs as base learners to perform dynamic self-training guided by both inter- and intra-model consistency as well as model error estimation. Specifically, we conduct a dynamic self-training process that comprises three-stage pseudo-label selection. In the first stage, we apply majority voting to generate ensembled pseudo-labels, ensuring reliable supervision. Then, we assess each model’s bias on a sample using weak and strong augmentations: predictions consistent under no and weak augmentation but changing under strong augmentation indicate reliability and less prone to pre-training bias. Finally, based on theoretical insights, we dynamically adjust the upper bound on the number of pseudo-labels used for training, according to models’ relative accuracy on labeled data. This further filters pseudo-labels and enables adaptive self-training throughout the training process.

As a plug-and-play approach, Bi-CoG can be seamlessly integrated into existing VLM fine-tuning pipelines. To evaluate its effectiveness, we conduct experiments under both \textit{open-world generalization} \cite{decoop} and \textit{standard semi-supervised learning}\cite{yang2022sslsurvey} settings, combining Bi-CoG with various prompt-tuning methods. Extensive results demonstrate the effectiveness of our bi-consistency-guided dynamic self-training strategy, which consistently improves performance across 14 datasets without relying on larger models or external knowledge, outperforming state-of-the-art (SOTA) methods.

Our main contributions are summarized as follows:

\begin{itemize}

    \item We propose Bi-CoG, a plug-and-play method that jointly leverages the pre-trained knowledge of VLMs and external information from unlabeled data.
    
    \item We design a deeply coupled self-training strategy that integrates inter-model consistency, intra-model consistency and model error estimation, enabling reliable, low-bias pseudo-label generation and adaptive training.
    \item Theoretical analysis and extensive experiments demonstrate that Bi-CoG consistently improves performance, achieving up to a 5.69\% gain across 14 datasets and surpassing SOTA methods.
\end{itemize}
\section{Related Work}

\paragraph{Parameter-efficient Fine-tuning on VLMs}

While pre-trained VLMs demonstrate strong zero-shot generalization in domain-specific downstream tasks, their large parameter sizes often hinder efficient adaptation to specific applications \cite{zhou2022coop}. To address this, recent works have proposed parameter-efficient fine-tuning approaches that introduce a small number of trainable parameters while keeping the majority of the model frozen \cite{vlm_peft_survey}. Tip-Adapter \cite{zhang2022tipadapter} and CLIP-Adapter \cite{gao2024clipadapter} append lightweight bottleneck layers after the text or image encoders to learn residual features for better representation. CoOp \cite{zhou2022coop} introduces prompt tuning to VLMs for the first time by replacing manually crafted hard prompts (e.g., “a photo of a {classname}”) with learnable soft text prompts optimized via backpropagation. Building on this idea, methods such as MaPLe \cite{khattak2023maple}, PromptSRC \cite{khattak2023promptsrc}, and BMIP \cite{lv2025bmip} extend prompt tuning to both visual and textual encoders, enabling joint vision-language adaptation. While these methods have advanced the application of VLMs to downstream tasks, the transformation of prompts into continuous vectors makes them prone to overfitting and leads to sub-optimal performance in open-world scenarios \cite{zhou2022cocoop,khattak2023promptsrc}. Beyond prompt tuning, recent work on VLM selection and reuse \cite{vmsr2025icml} further explores how to choose and adapt pre-trained models for downstream tasks.

\paragraph{Pseudo-labeling in Semi-supervised Learning} 
Semi-supervised learning (SSL) addresses label scarcity by leveraging both labeled and unlabeled data. Pseudo-labeling, the most widely used SSL approach, assigns labels to unlabeled data based on model predictions and improves performance through self-training.
The key to effective pseudo-labeling lies in the generation of high-quality pseudo-labels \cite{yang2022sslsurvey}. Traditional SSL methods typically filter pseudo-labels using confidence thresholds or adopt data augmentation to generate diverse views, aiming to improve label quality and model robustness~\cite{lee2013pseudolabel}. For VLMs, their strong generalization ability allows them to produce highly accurate pseudo-labels even in zero- or few-shot settings, making them strong candidates as pseudo-labelers. Recent works have explored more advanced strategies: GRIP~\cite{menghini2023grip} gradually increases the upper limit on the number of pseudo-labels to mitigate calibration errors~\cite{oh2024towards} and label imbalance~\cite{wang2022debiased,guo2022classimbalanced}; and CPL~\cite{zhang2024cpl} progressively refines pseudo-labels based on a confidence score matrix to avoid overconfidence.
Despite their effectiveness, these methods struggle with balancing pseudo-label noise and inherited pre-training biases~\cite{oh2024towards,wang2022debiased,li2021safeweakly}, while failing to dynamically adapt to model performance in open environments~\cite{guo2025robust}, ultimately limiting their potential.

\begin{figure*}[t]
\centering
\includegraphics[width=\textwidth]{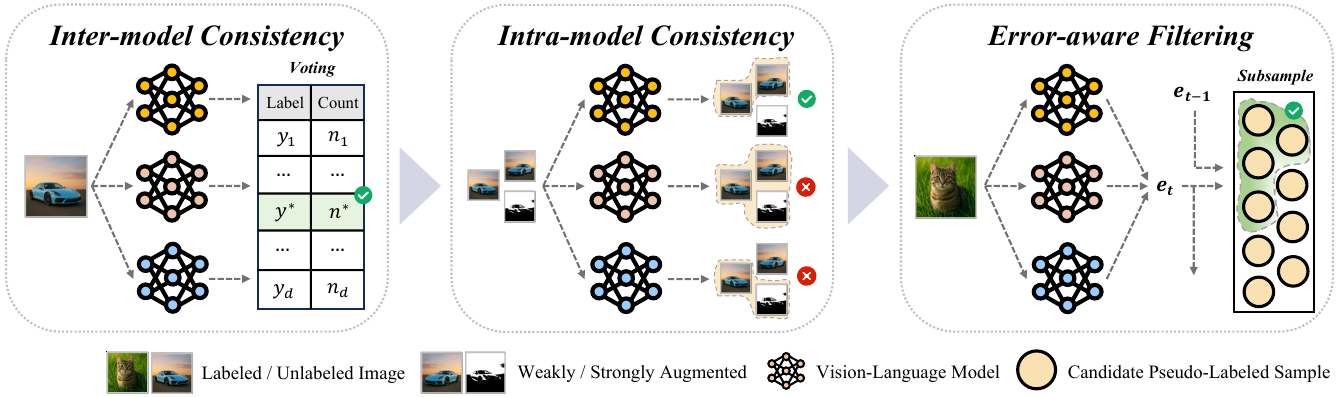}
\caption{The overall framework of Bi-CoG's pseudo-label selection. During the self-training phase, Bi-CoG employs inter-model consistency, intra-model consistency, and error-aware filtering to generate accurate and unbiased pseudo-labels.}
\label{pipeline}
\end{figure*}

\paragraph{Problem Definition}
Based on the aforementioned challenges in pseudo-labeling—specifically the issues of label noise and static adaptation—we formally define our task. In downstream scenarios with scarce labeled data, we typically have access to a small labeled dataset denoted as 
$D_L = \{(x_i, y_i)\}_{i=1}^M$, along with a large amount of unlabeled data 
$D_{UL} = \{x_i\}_{i=1}^N$, where $M \ll N$. 
Given a collection of $K$ pre-trained VLMs $\{f_k\}_{k=1}^K$ (with $K \geq 3$), 
the objective is to effectively leverage both $D_L$ and $D_{UL}$ to fine-tune the models and maximize performance on the target dataset $D_{test}$.
\section{Methodology}

To address the prevalent challenges of balancing pseudo-label accuracy with model bias and reducing the strong reliance on empirical parameter tuning in prior methods, we propose Bi-Consistency-Guided Self-Training (Bi-CoG), a simple yet effective plug-and-play framework.
As illustrated in Figure \ref{pipeline}, the Bi-CoG framework comprises the following three major components for pseudo-label selection:

\begin{itemize}
    \item A model ensembling module that aggregates predictions through majority voting based on inter-model consistency, generating high-quality candidate pseudo-labels to ensure reliable supervision.
    \item A multi-view filtering mechanism that leverages intra-model consistency by applying diverse transformations to unlabeled data and discarding biased samples with inconsistent predictions, thereby mitigating the influence of pre-training bias.
    \item An error-aware bounding strategy that dynamically estimates model performance and sets an upper limit on the number of pseudo-labels used, promoting stable and progressive self-training.
\end{itemize}

\begin{algorithm}[t]
\fontsize{9}{10}\selectfont
\caption{Bi-CoG's overall learning algorithm.}
\label{alg:algorithm}
\textbf{Input}: Pre-trained VLMs $\{f_{k}\}_{k=1}^K$, labeled data $D_{L}$, unlabeled data $D_{UL}$, scaling factor $\alpha$ \\
\textbf{Output}: Fine-tuned VLMs
\begin{algorithmic}[1] 
\FOR{$j = 1 \ \mathbf{to} \ K$}
    \STATE $e'_j \leftarrow 0.5$, $\mathrm{update}_j \leftarrow \mathrm{False}$
\ENDFOR
\STATE $t \leftarrow 0, \ \mathrm{improve} \leftarrow \mathrm{True}$
\WHILE{$\mathrm{improve}$}
    \STATE $t \leftarrow t + 1$
    \FOR{$j = 1 \ \mathbf{to} \ K$}
        \STATE $\mathrm{update}_j \leftarrow \mathrm{False}$
        \STATE $e_j \leftarrow \mathrm{MeasureError}(D_{L}, \{f_k\}_{\ k \neq j}^K)$
        \IF{$e_j < e_j' $}
            \STATE $D_{Inter}^j \leftarrow \mathrm{InterConsistency}(D_{UL}, \{f_k\}_{k \neq j}^K)$
            \STATE $D_{Intra}^j \leftarrow\mathrm{IntraConsistency}(D_{UL}, \{f_k\}_{k \neq j}^K)$
            \STATE $\mathrm{update}_j \leftarrow \mathrm{ErrorAware}\left(e_j, \ e_j', \ t\right)$ 
            \STATE $D_{PL}^j \leftarrow \mathrm{Subsample}(D_{Inter}^j \cap D_{Intra}^j,(\frac{e_j'}{e_j} )^{\alpha t} L_j')$ 
        \ENDIF
        \IF{$\mathrm{update}_j$}
            \STATE  $e_j' \leftarrow e_j$, \ $f_j \leftarrow \mathrm{FineTune}(f_j, D_{L} \cup D_{PL}^j)$
        \ENDIF
    \ENDFOR

    \IF{$\mathrm{update} == \left[\mathrm{False} \right] * K$}
        \STATE $\mathrm{improve} \leftarrow \mathrm{False} $
        
    \ENDIF
\ENDWHILE
\STATE \textbf{return} $\{f_{k}\}_{k=1}^K$
\end{algorithmic}
\end{algorithm}

The procedure of Bi-CoG is summarized in Algorithm \ref{alg:algorithm}. We elaborate on each component in the subsequent sections.

\subsection{Inter-model Consistency} 

During self-training with pseudo-labels, label fidelity is critical to effective and stable optimization \cite{learn-from-noisy-examples}. Excessive noise can mislead learning, causing convergence to suboptimal representations. To mitigate this issue, we first generate pseudo-labels based on inter-model consistency by integrating predictions of multiple VLMs, with the primary goal of ensuring the accuracy of pseudo-labels. 

Concretely, for each model $f_j$ ($1 \leq j \leq K$), we construct a candidate pseudo-labeled dataset $D_{Inter}^j$ by selecting unlabeled instances on which the majority of the other $K - 1$ models reach a consensus. For each unlabeled sample $x_i \in D_{UL}$, we first compute the class-wise vote count from the other models as follows:

\begin{equation}
    v_i(y) = \sum_{1 \le k \le K,\ k \neq j} \mathbb{I}\left[f_k(x_i) = y\right]
\end{equation}

If the most voted label $y^* = \arg\max_{y \in \mathcal{Y}} v_i(y)$ satisfies $v_i(y^*) \ge \left\lceil \frac{K - 1}{2} \right\rceil$, we assign $x_i$ a pseudo-label $\hat{y}_i = y^*$ and include the pair $(x_i, y^*)$ in $D_{Inter}^j$. Otherwise, the sample is discarded. Formally, this candidate set is defined as:

\begin{equation}
    D_{Inter}^j = \left\{(x_i, y^*) \ \middle| \ v_i(y^*) \ge \left\lceil \frac{K - 1}{2} \right\rceil \right\}
\end{equation}
where $\mathcal{Y}$ denotes the label space, and $\mathbb{I}[\cdot]$ is the indicator function.
This strategy ensembles multiple models to generate reliable pseudo-labels, reducing noise from individual prediction errors. Moreover, the leave-one-out scheme prevents amplifying biases inherent to the target model, thereby resulting in more accurate and unbiased data for fine-tuning.

\subsection{Intra-model Consistency}
While inter-model consistency improves pseudo-label reliability by aggregating agreement across models, it may still preserve instances where VLMs share similar biases, often due to overlapping pre-training corpora or architectural homogeneity~\cite{erasing-the-bias,wang2023overwriting}.
To further mitigate biased supervision, we introduce intra-model consistency, which leverages the consistency of a single model's predictions under different input views to refine the pseudo-labeled sets obtained in the previous stage.

Specifically, for each unlabeled sample $x_i \in D_{UL}$, we apply both weak and strong data augmentations. Each model $f_k$ then produces predictions on the original input $x_i$, the weakly augmented $\tilde{x}_i^{w}$, and the strongly augmented $\tilde{x}_i^{s}$. A sample is retained only if the prediction remains consistent between the original and weakly augmented inputs, i.e., $f_k(x_i) = f_k(\tilde{x}_i^{w})$, but differs under strong augmentation, i.e., $f_k(\tilde{x}_i^{s}) \neq f_k(x_i)$. This criterion identifies samples with low uncertainty and meaningful sensitivity, as evidenced by prediction consistency under weak perturbations and inconsistency under strong perturbations. The intra-model consistent pseudo-label set for $f_j$ is defined as the intersection of predictions from the other models:

\begin{equation}
D_{Intra}^j = \bigcap_{\substack{k=1 \\ k \neq j}}^K\left\{ (x_i, f_k(x_i)) \middle|
\begin{array}{l}
f_k(x_i) = f_k(\tilde{x}_i^{w})\\
f_k(x_i) \neq f_k(\tilde{x}_i^{s})
\end{array}
\right\}
\end{equation}

Once the $D_{Intra}^j$ are computed, the candidate pseudo-label set is constructed by intersecting its inter-model consistent set $D_{Inter}^j$ with the intra-model consistent set:

\begin{equation}
D_{PL}^j = D_{Inter}^j \cap D_{Intra}^j 
\end{equation}

The intersection of pseudo-labels from inter-model consistency and intra-model consistency preserves accuracy while mitigating shared model biases. Consequently, the resulting pseudo-labeled datasets are both highly reliable and low in bias, providing a robust foundation for the subsequent dynamic self-training stage.

\subsection{Error-aware Filtering}

After bi-consistency-guided pseudo-labeling, we obtain a candidate sample set $D_{PL}$, which contains high-accuracy and low-biased pseudo-labels. However, during self-training, it is crucial to determine an appropriate upper bound on the number of pseudo-labeled samples based on model performance. Prior studies~\cite{learn-from-noisy-examples,debiased_self_training} have shown that even minor label noise, if not aligned with model competence, can significantly degrade performance, making further noise reduction in $D_{PL}$ essential. Existing methods~\cite{menghini2023grip,huang2022upl} typically rely on fixed thresholds to limit the number of pseudo-labels, which can be suboptimal as model performance evolves over time and across different datasets.

To address this limitation, we dynamically adjust the upper bound on pseudo-label usage based on estimated model error ratio, guided by a theoretical analysis. Specifically, following prior work \cite{learn-from-noisy-examples,tritraining} we formalize the influence of noisy supervision on model $f_j$ during iterative self-training:

\paragraph{Lemma 1.}
\textit{Let $L_t$ denote the training set $D_{train}$ for model $f_j$ at iteration $t$, and $e_t$ represent the relative error rate of the remaining $K - 1$ models at the same iteration, i.e., the probability that the majority vote produces an incorrect label. Then, training $f_j$ on $D_{train}$ will result in performance improvement if the following condition is satisfied:}

\begin{equation}
\label{equation:origin_condition}
0 < \frac{e_t}{e_{t-1}} < \frac{L_{t-1}}{L_t} < 1
\end{equation}

According to Lemma 1, the model will be improved throughout the training process as long as Equation \ref{equation:origin_condition} holds. 
\begin{table*}[t]
    \centering
    \setlength{\tabcolsep}{0.6mm}  
    \fontsize{9}{12}\selectfont   
    \begin{tabular}{l|cc|cc|cc|cc|cc}
    \toprule
    & \multicolumn{2}{c|}{Average}  & \multicolumn{2}{c|}{ImageNet}  & \multicolumn{2}{c|}{Caltech101} & \multicolumn{2}{c|}{OxfordPets} & \multicolumn{2}{c}{StanfordCars} \\  
    Method
    & \multicolumn{1}{c}{HM}      & \multicolumn{1}{c|}{Acc.}  & \multicolumn{1}{c}{HM}      & \multicolumn{1}{c|}{Acc.}  & \multicolumn{1}{c}{HM}      & \multicolumn{1}{c|}{Acc.}  & \multicolumn{1}{c}{HM}      & \multicolumn{1}{c|}{Acc.} & \multicolumn{1}{c}{HM}      & \multicolumn{1}{c}{Acc.} \\ 
    \midrule 
    CLIP & 73.00 & 66.41 & 70.18 & 66.70 & 95.68 & 93.30 & 94.11 & 89.10 & 68.85 & 65.60 \\  
    \midrule
    CoOp & 72.10 & 66.59 & 72.00$_{\pm 0.34}$ & 68.73$_{\pm 0.25}$ & 90.91$_{\pm 1.61}$ & 89.90$_{\pm 0.99}$ & 93.97$_{\pm 0.72}$ & 88.73$_{\pm 0.09}$ & 69.94$_{\pm 0.59}$ & 66.40$_{\pm 0.50}$ \\  
    \rowcolor{gray!20}
    \textit{+ Bi-CoG} & \textbf{77.79} & \textbf{71.03} & \textbf{73.06}$_{\pm 0.08}$ & \textbf{69.78}$_{\pm 0.04}$ & \textbf{96.67}$_{\pm 0.05}$ & \textbf{94.24}$_{\pm 0.25}$ & \textbf{96.28}$_{\pm 0.60}$ & \textbf{89.57}$_{\pm 2.02}$ & \textbf{73.38}$_{\pm 0.31}$ & \textbf{69.31}$_{\pm 0.76}$ \\  
    \midrule
    MaPLe & 78.78 & 72.18 & 73.58$_{\pm 0.09}$ & 70.30$_{\pm 0.14}$ & 96.58$_{\pm 0.21}$ & 94.87$_{\pm 0.09}$ & 96.60$_{\pm 0.36}$ & 92.33$_{\pm 0.31}$ & 73.53$_{\pm 0.81}$ & 69.97$_{\pm 0.87}$ \\  
    \rowcolor{gray!20}
    \textit{+ Bi-CoG} & \textbf{79.77} & \textbf{73.54} & \textbf{74.11}$_{\pm 0.07}$ & \textbf{70.82}$_{\pm 0.09}$ & \textbf{96.77}$_{\pm 0.17}$ & \textbf{95.09}$_{\pm 0.46}$ & \textbf{96.93}$_{\pm 0.06}$ & \textbf{92.56}$_{\pm 0.12}$ & \textbf{74.20}$_{\pm 0.31}$ & \textbf{70.48}$_{\pm 0.36}$ \\  
    \midrule
    PromptSRC & 79.99 & 74.11 & 74.07$_{\pm 0.06}$ & 70.77$_{\pm 0.05}$ & 96.11$_{\pm 0.08}$ & 94.63$_{\pm 0.12}$ & 96.26$_{\pm 0.29}$ & 92.13$_{\pm 0.59}$ & 76.68$_{\pm 0.16}$ & 73.03$_{\pm 0.19}$ \\  
    \rowcolor{gray!20}
    \textit{+ Bi-CoG} & \textbf{81.46} & \textbf{74.96} & \textbf{74.18}$_{
    \pm 0.03}$ & \textbf{70.94}$_{
    \pm 0.03}$ & \textbf{96.45}$_{\pm 0.16}$ & \textbf{94.74}$_{\pm 0.16}$ & \textbf{96.37}$_{\pm 0.13}$ & \textbf{92.62}$_{\pm 0.13}$ & \textbf{79.07}$_{\pm 0.15}$ & \textbf{73.77}$_{\pm 0.01}$ \\
    \midrule
    \midrule
     & \multicolumn{2}{c|}{Flowers102} & \multicolumn{2}{c|}{Food101} & \multicolumn{2}{c|}{FGVCAircraft} & \multicolumn{2}{c}{SUN397} & \multicolumn{2}{c}{DTD}\\  
    Method & \multicolumn{1}{c}{HM}      & \multicolumn{1}{c|}{Acc.}  & \multicolumn{1}{c}{HM} & \multicolumn{1}{c|}{Acc.}  & \multicolumn{1}{c}{HM}      & \multicolumn{1}{c|}{Acc.}  & \multicolumn{1}{c}{HM}      & \multicolumn{1}{c|}{Acc.} & \multicolumn{1}{c}{HM}      & \multicolumn{1}{c}{Acc.} \\  
    \midrule 
    CLIP & 74.44 & 70.70 & 90.39 & 85.60 & 31.21 & 24.80 & 72.37 & 62.60 & 56.62 & 44.00 \\  
    \midrule
    CoOp & 75.70$_{\pm 2.82}$ & 71.27$_{\pm 2.21}$ & 85.82$_{\pm 1.41}$ & 80.03$_{\pm 1.17}$ & 29.93$_{\pm 1.54}$ & 26.07$_{\pm 0.74}$ & 71.60$_{\pm 1.81}$ & 63.33$_{\pm 1.44}$ & 55.93$_{\pm 1.25}$ & 50.87$_{\pm 0.62}$ \\  
    \rowcolor{gray!20}
    \textit{+ Bi-CoG} & \textbf{83.57}$_{\pm 0.91}$ & \textbf{79.59}$_{\pm 1.22}$ & \textbf{90.54}$_{\pm 0.14}$ & \textbf{85.03}$_{\pm 0.65}$ & \textbf{35.61}$_{\pm 0.29}$ & \textbf{28.40}$_{\pm 0.21}$ & \textbf{76.89}$_{\pm 0.06}$ & \textbf{67.15}$_{\pm 0.18}$ & \textbf{59.90}$_{\pm 1.37}$ & \textbf{50.96}$_{\pm 1.35}$ \\  
    \midrule
    MaPLe & 82.98$_{\pm 0.19}$ & 78.47$_{\pm 0.60}$ & 90.83$_{\pm 0.35}$ & 86.47$_{\pm 0.40}$ & 35.30$_{\pm 0.88}$ & 27.30$_{\pm 0.96}$ & 79.42$_{\pm 0.30}$ & 70.97$_{\pm 0.24}$ & 67.40$_{\pm 3.09}$ & 55.30$_{\pm 1.40}$ \\  
    \rowcolor{gray!20}
    \textit{+ Bi-CoG} & \textbf{84.62}$_{\pm 0.25}$ & \textbf{79.12}$_{\pm 0.51}$ & \textbf{91.67}$_{\pm 0.03}$ & \textbf{87.46}$_{\pm 0.02}$ & \textbf{35.58}$_{\pm 0.71}$ & \textbf{27.63}$_{\pm 0.56}$ & \textbf{80.27}$_{\pm 0.19}$ & \textbf{71.82}$_{\pm 0.23}$ & \textbf{69.88}$_{\pm 1.07}$ & \textbf{56.36}$_{\pm 0.73}$ \\  
    \midrule
    PromptSRC & 86.13$_{\pm 0.39}$ & 80.77$_{\pm 0.50}$ & 90.91$_{\pm 0.12}$ & 86.47$_{\pm 0.12}$ & 34.99$_{\pm 6.68}$ & 29.93$_{\pm 2.50}$ & 80.68$_{\pm 0.23}$ & 72.13$_{\pm 0.26}$ & 70.72$_{\pm 1.23}$ & 58.87$_{\pm 1.28}$ \\  
    \rowcolor{gray!20}
    \textit{+ Bi-CoG} & \textbf{87.60}$_{\pm 0.18}$ & \textbf{83.19}$_{\pm 0.55}$ & \textbf{91.25}$_{\pm 0.01}$ & \textbf{86.65}$_{\pm 0.02}$ & \textbf{39.89}$_{\pm 0.34}$ & \textbf{31.98}$_{\pm 0.69}$ & \textbf{80.89}$_{\pm 0.04}$ & \textbf{72.43}$_{\pm 0.10}$ & \textbf{71.67}$_{\pm 0.42}$ & \textbf{59.48}$_{\pm 0.10}$   \\
    \midrule
    \midrule
    & \multicolumn{2}{c|}{EuroSAT} & \multicolumn{2}{c|}{UCF101} & \multicolumn{2}{c}{CIFAR-10} & \multicolumn{2}{c}{CIFAR-100} & \multicolumn{2}{c}{ImageNet-100} \\  
    Method & \multicolumn{1}{c}{HM}      & \multicolumn{1}{c|}{Acc.}  & \multicolumn{1}{c}{HM}      & \multicolumn{1}{c|}{Acc.}  & \multicolumn{1}{c}{HM}      & \multicolumn{1}{c|}{Acc.}  & \multicolumn{1}{c}{HM}      & \multicolumn{1}{c|}{Acc.} & \multicolumn{1}{c}{HM}      & \multicolumn{1}{c}{Acc.} \\  
    \midrule 
    CLIP & 60.21 & 48.30 & 74.42 & 67.50 & 88.43 & 79.00 & 55.17 & 46.00 & 88.33 & 86.50 \\  
    \midrule
    CoOp & 72.04$_{\pm 3.29}$ & 59.50$_{\pm 2.10}$ & 67.03$_{\pm 1.48}$ & 64.23$_{\pm 1.26}$ & 89.17$_{\pm 1.32}$ & 77.00$_{\pm 2.42}$ & 53.42$_{\pm 3.23}$ & 45.03$_{\pm 2.88}$ & 81.97$_{\pm 2.38}$ & 81.17$_{\pm 1.96}$ \\  
    \rowcolor{gray!20}
    \textit{+ Bi-CoG} & \textbf{85.84}$_{\pm 1.98}$ & \textbf{67.98}$_{\pm 0.67}$ & \textbf{73.11}$_{\pm 1.99}$ & \textbf{68.01}$_{\pm 0.92}$ & \textbf{92.07}$_{\pm 0.46}$ & \textbf{82.28}$_{\pm 0.99}$ & \textbf{61.71}$_{\pm 0.26}$ & \textbf{52.55}$_{\pm 0.32}$ & \textbf{90.39}$_{\pm 0.35}$ & \textbf{89.59}$_{\pm 0.30}$ \\  
    \midrule
    MaPLe & 78.35$_{\pm 1.46}$ & 60.90$_{\pm 8.71}$ & 81.57$_{\pm 0.39}$ & 74.70$_{\pm 0.57}$ & 92.02$_{\pm 0.76}$ & 83.83$_{\pm 1.21}$ & 64.73$_{\pm 0.30}$ & 56.10$_{\pm 0.37}$ & 90.00$_{\pm 0.17}$ & 89.07$_{\pm 0.33}$ \\  
    \rowcolor{gray!20}
    \textit{+ Bi-CoG} & \textbf{81.33}$_{\pm 2.89}$ & \textbf{70.52}$_{\pm 2.81}$ & \textbf{82.23}$_{\pm 0.58}$ & \textbf{75.75}$_{\pm 0.51}$ & \textbf{92.15}$_{\pm 0.09}$ & \textbf{84.41}$_{\pm 0.69}$ & \textbf{66.68}$_{\pm 0.52}$ & \textbf{57.97}$_{\pm 0.61}$ & \textbf{90.41}$_{\pm 0.29}$ & \textbf{89.52}$_{\pm 0.32}$ \\  
    \midrule
    PromptSRC & 81.87$_{\pm 2.52}$ & 70.03$_{\pm 0.86}$ & 82.30$_{\pm 0.87}$ & 76.67$_{\pm 0.37}$ & 91.46$_{\pm 0.30}$ & 83.80$_{\pm 0.42}$ & 67.23$_{\pm 0.42}$ & 58.93$_{\pm 0.39}$ & 90.49$_{\pm 0.22}$ & 89.43$_{\pm 0.26}$ \\  
    \rowcolor{gray!20}
    \textit{+ Bi-CoG} & \textbf{86.28}$_{\pm 0.87}$ & \textbf{71.01}$_{\pm 0.47}$ & \textbf{83.09}$_{\pm 0.24}$ & \textbf{77.30}$_{\pm 0.44}$ & \textbf{94.50}$_{\pm 0.15}$ & \textbf{84.96}$_{\pm 1.75}$ & \textbf{68.19}$_{\pm 0.38}$ & \textbf{60.20}$_{\pm 0.07}$ & \textbf{91.07}$_{\pm 0.02}$ & \textbf{90.15}$_{\pm 0.05}$ \\
    \bottomrule
    \end{tabular}
    \caption{Performance comparison on 14 datasets using ViT-B/16 architecture. The best performance is in bold.}  
    \label{tab:ow generalization}
\end{table*}
In practice, we estimate the theoretical error rate ratio from the performance of the $K-1$ models on the labeled set. However, this estimation can be biased, as repeated training on labeled data tends to produce stable—and often low—error ratios, which are reflected in Equation \ref{equation:origin_condition} as values close to 1. In contrast, the actual error ratio on unlabeled data typically decreases during training, making the true ratio \(\frac{e_t}{e_{t-1}}\) lower than the estimated \(\frac{\hat{e}_t}{\hat{e}_{t-1}}\). Ignoring this discrepancy would introduce bias into the estimated pseudo-label budget, compromising the validity of the theoretical analysis. To address this issue, we introduce Theorem 1 to provide a correction:

\paragraph{Theorem 1.}

\textit{As the training progresses, the true error rate ratio can be approximately equal to the \(t\)-power of the estimated error rate ratio, as shown below:}

\begin{equation}
\label{approx_condition}
\frac{e_t}{e_{t-1}} \approx \left[\frac{\hat{e}_t}{\hat{e}_{t-1}} \right]^{\alpha t}
\end{equation}
\textit{where $\alpha > 0$ is a scaling factor that reflects the error reduction rate. Then, the original condition in Equation \ref{equation:origin_condition} is approximately equivalent to the following constraints:}

\begin{equation}
\label{updated_condition}
\begin{aligned}
L_t &\leq \left\lceil \left(\frac{\hat{e}_{t-1}}{\hat{e}_t} \right)^{\alpha t} \times L_{t-1} - 1 \right\rceil \\
L_{t-1} &> \frac{\hat{e}_t^{\alpha t}}{\hat{e}_{t-1}^{\alpha t} - \hat{e}_t^{\alpha t}}
\end{aligned}
\end{equation}

Based on the theoretical analysis, we establish the necessary condition under which the model can be effectively optimized by self-training, as described in Equation \ref{updated_condition}, and establish an upper bound on the pseudo-label budget for each iteration. Here, $L_j'$ denotes the size of the pseudo-labeled training set used by model $f_j$ in the previous iteration. Accordingly, the final set of pseudo-labeled samples is formalized as:

\begin{equation}
    D_{PL^*}^j =  \mathrm{Subsample}\left(D_{PL}^j, \ \left(\frac{e_j'}{e_j}\right)^{\alpha t} L_j' \right)
\end{equation}

This dynamic filtering strategy reduces the mismatch between model capacity and pseudo-label quantity. Consequently, the model can self-regulate the amount of pseudo-labeled data during training, leading to stable and consistent performance improvements.

After training, Bi-CoG adopts a majority voting strategy to generate predicted labels for evaluation during the testing phase, ensuring robust and reliable assessment results. 

In summary, Bi-CoG offers the following key advantages:
\begin{enumerate}
    \item \textbf{Plug-and-Play Adaptability}: Bi-CoG can be seamlessly integrated with existing pre-trained models, providing consistent performance improvements with minimal changes to the original method.
    \item \textbf{Robust Pseudo-Labeling}: By jointly leveraging bi-consistency and error-aware mechanisms, Bi-CoG generates more reliable pseudo-labels, effectively reducing label noise and alleviating model bias.
    \item \textbf{Adaptive Self-Training Control}: Empowered by the error-aware mechanism, Bi-CoG can automatically regulate the self-training process, eliminating the need for manually preset iteration counts and enhances adaptability across different scenarios.
\end{enumerate}
\section{Experiments}
In this section, we design experiments to answer the following research questions:
\begin{enumerate}
    \item \textbf{RQ1:} Does Bi-CoG enhance existing PEFT methods for VLMs in open-world generalization scenarios?
    \item \textbf{RQ2:} Can Bi-CoG deliver competitive performance compared to conventional SSL approaches?
    \item \textbf{RQ3:} How does Bi-CoG compare to other methods that jointly leverage pre-trained models and unlabeled data?
    \item  \textbf{RQ4:} What is the contribution of each component within the Bi-CoG framework?
\end{enumerate}

\subsection{Experimental Settings}
\paragraph{Evaluation Paradigm.} To answer RQ1 and RQ2, we evaluate the effectiveness of Bi-CoG on two widely adopted settings: \emph{open-world generalization setting}~\cite{decoop,lv2025bmip} and \emph{semi-supervised setting}~\cite{lee2013pseudolabel,yang2022sslsurvey}.
In \emph{open-world generalization setting}, the model is trained only on a subset of categories (referred to as base classes), while the remaining categories are treated as novel classes. We report the harmonic mean (HM) of the model’s performance on base and novel classes, as well as the overall accuracy on the full test set, which includes both base and novel classes. \emph{Semi-supervised setting} follows a common protocol where only a small portion of the data is labeled, while the majority remains unlabeled.

\paragraph{Dataset and Baselines.} We evaluate the performance of our method on 14 recognition datasets, including 11 commonly used for assessing VLMs and 3 classical benchmarks for semi-supervised learning. We select CLIP~\cite{radford2021clip} and its three representative prompt tuning methods as baselines: CoOp~\cite{zhou2022coop}, MaPLe~\cite{khattak2023maple}, and PromptSRC~\cite{khattak2023promptsrc}, among which PromptSRC represents the current SOTA method. For methods that integrate the two paradigms, we select GRIP~\cite{menghini2023grip} and CPL~\cite{zhang2024cpl} as baselines, with CPL representing the current SOTA approach.

\paragraph{Implementation Details.}
We set the number of VLMs $K = 3$ throughout all experiments. The setting $K \ge 3$ ensures that an absolute majority can be achieved in most cases, facilitating reliable pseudo-label selection. While a larger $K$ further improves stability, it also increases GPU memory consumption. We adopt $K = 3$ to balance effectiveness and efficiency. For the scaling factor $\alpha$, it provides a controllable mechanism to adjust the strictness of pseudo-label filtering: a smaller $\alpha$ enhances the utilization of more pseudo-labels, while a larger $\alpha$ suppresses noisy pseudo-labels by raising the filtering threshold. We uniformly set $\alpha = 1$ across all datasets to ensure generalizability without dataset-specific tuning. All experiments are conducted on a single NVIDIA A800 GPU.

\subsection{Experimental Results}

\paragraph{RQ1:} Does Bi-CoG enhance existing PEFT methods for VLMs in open-world generalization scenarios?

To evaluate the robustness of Bi-CoG in open-world scenarios, we conduct a comprehensive study by integrating it with three representative prompt tuning methods across 14 datasets. Notably, in all experiments related to Bi-CoG, we controlled the random seeds during both the initialization and training phases for different models to ensure diversity in the initial training stages. For each dataset, we report two metrics: HM of base and novel class accuracies, and overall accuracy (Acc.). We set the base-to-novel class ratio to 50:50 and report the mean and standard deviation over three runs. As shown in Table~\ref{tab:ow generalization}, Bi-CoG consistently improves both metrics across all datasets, achieving the highest average performance. Notably, when applied to CoOp, Bi-CoG yields over a 5\% improvement in HM. Moreover, Bi-CoG demonstrates particularly strong gains on traditional open-world benchmarks. For example, on low-resolution datasets such as CIFAR-10 and CIFAR-100 \cite{cifar}, it achieves up to an 8\% increase in HM, underscoring its effectiveness across datasets with varying resolutions. 
These results highlight the effectiveness of Bi-CoG and emphasize the importance of integrating unsupervised strategies with pre-trained models in open-world settings.

\paragraph{RQ2:} Can Bi-CoG deliver competitive performance compared to conventional SSL approaches?

\begin{table}
\centering
    \fontsize{9}{10}\selectfont
    \setlength{\tabcolsep}{.5mm}
    \begin{tabular}{lcccc}
    \toprule
    \multirow{2}{*}{Method} & \multicolumn{2}{c}{CIFAR-10} & \multicolumn{2}{c}{CIFAR-100} \\
    \cmidrule(lr){2-3} \cmidrule(lr){4-5}
    & 4 shots & 25 shots & 4 shots & 25 shots  \\
    \midrule
    ReMixMatch     & 80.90$_{\pm 9.64}$  & 94.56$_{\pm 0.05}$   & 55.72$_{\pm 2.06}$ & \textbf{72.57}$_{\pm 0.31}$ \\
    FixMatch (RA)  & 86.19$_{\pm 3.37}$  & 94.93$_{\pm 0.65}$   & 51.15$_{\pm 1.75}$ & 71.71$_{\pm 0.11}$ \\
    FixMatch (CTA) & 88.61$_{\pm 3.35}$ & \textbf{94.93}$_{\pm 0.33}$   & 50.05$_{\pm 3.01}$ & 71.36$_{\pm 0.24}$ \\ 
    
    \midrule
    CoOp & 80.60$_{\pm 1.79}$ & 85.10$_{\pm 0.29}$ & 49.00$_{\pm 0.83}$ & 57.43$_{\pm 0.25}$ \\
    \rowcolor{gray!20}
    \textit{+ Bi-CoG} & 84.90$_{\pm 1.06}$ & 86.10$_{\pm 0.03}$ & 52.77$_{\pm 0.24}$ & 58.05$_{\pm 0.05}$ \\ 
    
    \midrule
    PromptSRC & 86.03$_{\pm 0.48}$ & 90.10$_{\pm 0.08}$ & 60.33$_{\pm 0.17}$ & 66.33$_{\pm 0.29}$ \\
    \rowcolor{gray!20}
    \textit{+ Bi-CoG} & \textbf{89.83}$_{\pm 0.35}$ & 91.07$_{\pm 0.42}$ & \textbf{60.56}$_{\pm 0.79}$ & 66.84$_{\pm 0.67}$ \\ 
    
    \bottomrule
    \end{tabular}
    \caption{Performance on low-resolution SSL datasets.}  
    \label{tab:ssl}
\end{table}

As shown in Table~\ref{tab:ssl}, SSL algorithms exhibit a significant advantage over baseline prompt tuning methods on SSL datasets. This is largely due to the lower resolution of these datasets, which makes it more difficult for VLMs to effectively capture visual content~\cite{cifar,unlabel_or_pretrain}. Notably, this advantage also comes at the cost of substantial computational overhead: conventional SSL methods typically require over 10 hours of training for the CIFAR dataset, whereas Bi-CoG achieves comparable or superior performance with only 1 hour of training under the same hardware configuration.
This limitation of high computational cost, coupled with the sensitivity to the number of shots (increasing the number of shots from 4 to 25 yields much larger performance gains for SSL algorithms compared to VLM-based prompt tuning), highlights the practical challenges of traditional SSL methods. In contrast, Bi-CoG provides the model with more reliable training examples via adaptive pseudo-labeling, enabling prompt tuning methods to achieve SOTA performance under the challenging 4-shot setting while maintaining remarkable computational efficiency.

These results demonstrate that Bi-CoG can effectively leverage unlabeled data to capture richer visual representations even in low-resolution and data-scarce scenarios, while also significantly reducing training time compared to SSL methods, making it more practical and scalable.

\paragraph{RQ3:} How does Bi-CoG compare to other methods that jointly leverage pre-trained models and unlabeled data?

To further validate the effectiveness of Bi-CoG, we conducted comparative experiments against two SOTA methods: GRIP \cite{menghini2023grip} and CPL \cite{zhang2024cpl}. Specifically, GRIP defines a fixed epoch-dependent expression for selecting the number of pseudo-labels, while CPL relies on manually set confidence thresholds to filter pseudo-labels iteratively. Following the experimental protocol of GRIP, we adopted the same base-to-novel class ratio (62:38) and used CoOp as the underlying prompt tuning framework for Bi-CoG and all compared methods. This ensures a fair comparison by eliminating discrepancies from different base models or tuning strategies. As shown in Table~\ref{tab:GRIP-compare}, Bi-CoG achieves the best performance under both SSL and open-world settings, outperforming CPL by up to 5.9 percentage points. Notably, this superiority stems from Bi-CoG’s adaptive pseudo-label selection strategy during iterations: unlike GRIP’s epoch-dependent rule or CPL’s manual threshold tuning, Bi-CoG adjusts pseudo-label selection based on model consistency and estimated error rate, ensuring high-quality supervision signals. More importantly, Bi-CoG dispenses with the requirement for manually tuning hyperparameters, effectively reducing dependence on human expertise. In summary, Bi-CoG represents one of the most effective integrations of pretrain-finetuning and semi-supervised learning methods to date.

\subsection{Ablation Study}

\begin{figure}[t]
\centering
\includegraphics[width=\columnwidth]{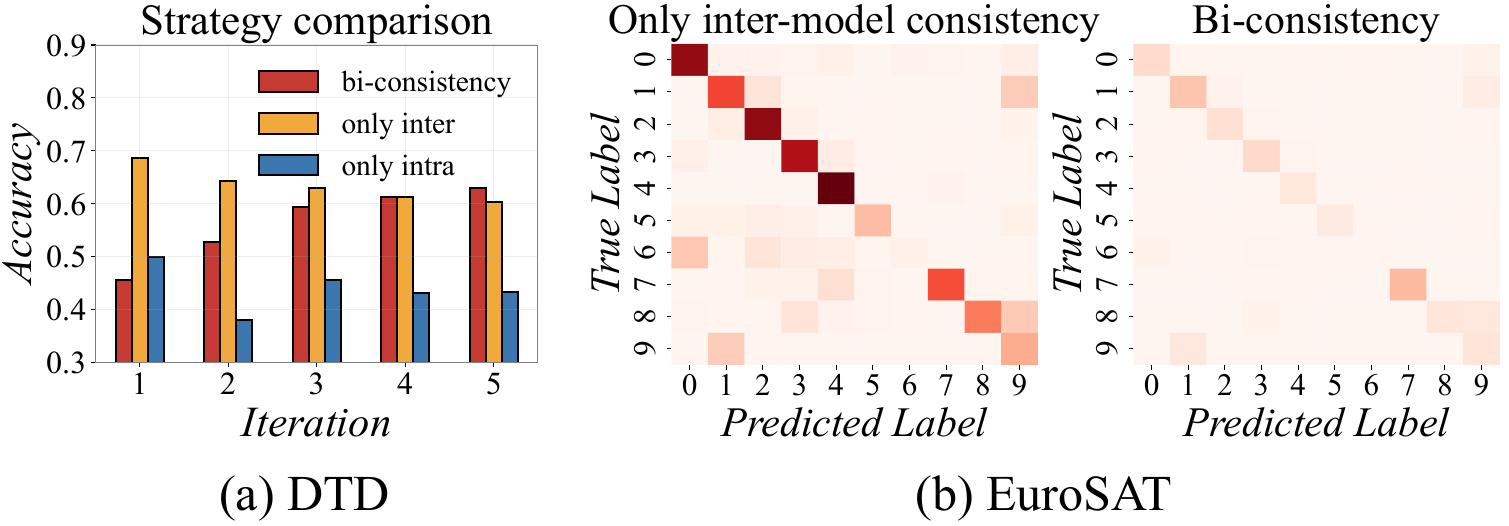} 
\caption{(a) Pseudo-label accuracy on DTD under different pseudo-labeling strategies; (b) Distribution of pseudo-labels on EuroSAT before and after applying the intra-model consistency.}
\label{ablation_1}
\end{figure}

To answer \textbf{RQ4}, we first examine the impact of the bi-consistency-guided pseudo-labeling strategy on both pseudo-label accuracy and the distribution of biased samples. As shown in Figure \ref{ablation_1}.a, removing the inter-model consistency module results in a notable decline in pseudo-label accuracy, highlighting the importance of ensemble-based labeling for generating high-quality supervision. Notably, although the bi-consistency strategy exhibits relatively lower pseudo-label accuracy in the early stages of training, its accuracy steadily improves as training progresses. In contrast, other single-strategy pseudo-labeling methods show either a decline or fluctuation in accuracy, demonstrating the stability and reliability of bi-consistency-guided pseudo-labeling. Furthermore, as illustrated in Figure \ref{ablation_1}.b, incorporating intra-model consistency helps smooth the class-wise distribution of pseudo-labels, effectively mitigating the bias inherited from pre-training. These results collectively confirm that the bi-consistency-guided strategy produces pseudo-labels that are both more accurate and less biased—two critical factors for stable and effective self-training.

\begin{figure}[t]
\centering
\includegraphics[width=0.73\columnwidth]{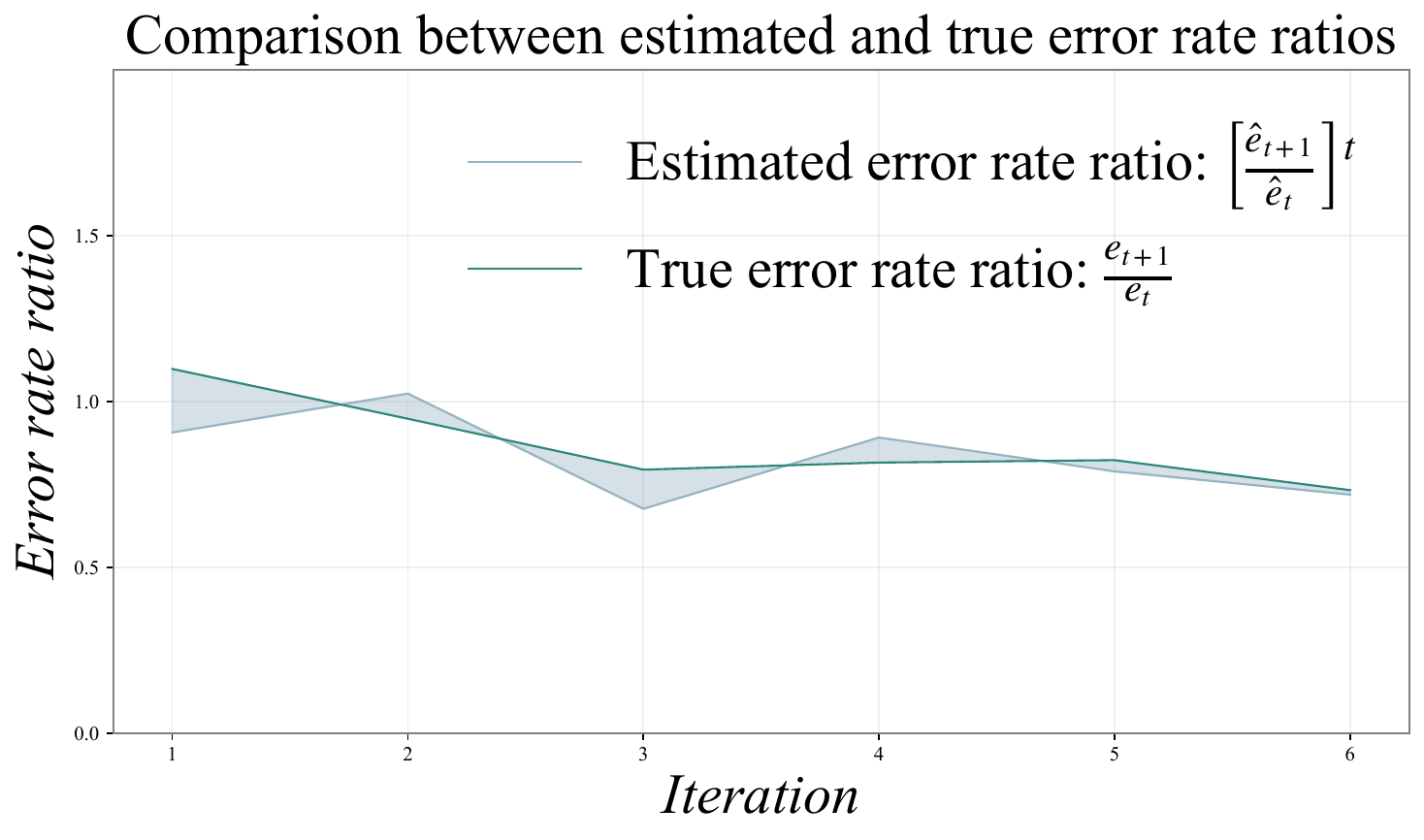} 
\caption{Comparison between the estimated and true error rate ratios. The scaling factor is fixed at $\alpha = 1$ throughout training.}
\label{fig:ablation_2}
\end{figure}

\begin{table}[t]
\centering
\fontsize{9}{10}\selectfont
\begin{tabular}{cccc}
\toprule
Method & Domain Data & SSL & Open-World  \\
\midrule
CLIP & Zero-shot & 72.08 & 77.80 \\
\midrule
FPL & \multirow{5}{*}{\shortstack{Few-shot \\ + \\ Unlabeled}} & 75.96 & 80.97 \\
IFPL & & 78.68 & 82.08 \\
GRIP & & 83.60 & 86.26 \\
CPL & & 89.66 & 87.35 \\
\cellcolor{gray!15}Bi-CoG  & & \cellcolor{gray!15}\textbf{96.45} & \cellcolor{gray!15}\textbf{88.01}  \\
\midrule
$\Delta$ &  & +6.79 & +0.66 \\
\bottomrule
\end{tabular}
\caption{Performance comparison on the Flowers102 dataset.}
\label{tab:GRIP-compare}
\end{table}

\begin{table}[t]
\centering
    \fontsize{9}{12}\selectfont
    \begin{tabular}{lcc}
    \toprule
    \multirow{2}{*}{Method} & \multicolumn{2}{c}{Open-World} \\
    \cmidrule(lr){2-3}
    & HM & Acc. \\
    \midrule
    \rowcolor{gray!20}
    CoOp + Bi-CoG & 77.79 & 71.03  \\
    \textit{- w/o. Inter-model consistency} & 76.14 & 68.83 \\ 
    \textit{- w/o. Intra-model consistency} & 75.85 & 69.08 \\
    \textit{- w/o. Error-aware filtering} & 76.92 & 70.92 \\
    
    \bottomrule
    \end{tabular}
    \caption{Ablation study of Bi-CoG's pseudo-labeling strategy.}  
    \label{tab:ablation}
\end{table}

To further validate the theoretical foundation of our error-aware mechanism, we analyze the evolution of the actual error rate ratio on pseudo-labeled samples and its estimation from labeled data on StanfordCars \cite{stanfordcars} under the open-world generalization setting. As illustrated in Figure \ref{fig:ablation_2}, the condition in Equation \ref{approx_condition} holds reliably during training, supporting the practical validity of our theoretical derivation and confirming the soundness of Bi-CoG.

Finally, we conduct a comprehensive ablation study on
key components of Bi-CoG, using CoOp as the base model and reporting the average performance across 14 datasets. As shown in Table \ref{tab:ablation}, any component of our three-stage pseudo-labeling strategy, which integrates bi-consistency and error-aware filtering, plays a critical role in improving model performance. This highlights the deeply coupled and effective design of the pseudo-labeling mechanism in Bi-CoG.
\section{Conclusion}

This paper investigates the integration of pre-trained VLMs with semi-supervised learning for downstream tasks with sparse labeled data. We identify critical issues in existing approaches, including the trade-off between label quality and model bias, as well as the limited adaptability arising from the reliance on manually set fixed thresholds. To address these challenges, we propose Bi-CoG, a plug-and-play self-training method that incorporates a deeply coupled pseudo-labeling strategy with error-aware dynamic filtering capabilities. Theoretical analysis and experimental results across 14 datasets in both open-world generalization and semi-supervised settings demonstrate that Bi-CoG can be seamlessly integrated with any pretrain-finetuning methodologies, consistently achieving performance improvements.

\appendix

\section*{Acknowledgments}

This research was supported by the Key Program of Jiangsu Science Foundation (BK20243012), the National Science Foundation of China (62306133), and the ``111 Center'' (No. B26023).

\bibliographystyle{named}
\bibliography{ijcai26}

@inproceedings{radford2021clip,
  title={Learning Transferable Visual Models from Natural Language Supervision},
  author={Radford, Alec and Kim, Jong Wook and Hallacy, Chris and Ramesh, Aditya and Goh, Gabriel and Agarwal, Sandhini and Sastry, Girish and Askell, Amanda and Mishkin, Pamela and Clark, Jack and others},
  booktitle={Proceedings of the 38th International Conference on Machine Learning},
  pages={8748--8763},
  year={2021}
}

@inproceedings{wang2023overwriting,
  title={Overwriting Pretrained Bias with Finetuning Data},
  author={Wang, Angelina and Russakovsky, Olga},
  booktitle={Proceedings of the IEEE/CVF International Conference on Computer Vision},
  pages={3957--3968},
  year={2023}
}

@inproceedings{jia2021align,
  title={Scaling Up Visual and Vision-Language Representation Learning with Noisy Text Supervision},
  author={Jia, Chao and Yang, Yinfei and Xia, Ye and Chen, Yi-Ting and Parekh, Zarana and Pham, Hieu and Le, Quoc and Sung, Yun-Hsuan and Li, Zhen and Duerig, Tom},
  booktitle={Proceedings of the 38th International Conference on Machine Learning},
  pages={4904--4916},
  year={2021}
}

@article{gao2024clipadapter,
  title={{CLIP}-Adapter: Better Vision-Language Models with Feature Adapters},
  author={Gao, Peng and Geng, Shijie and Zhang, Renrui and Ma, Teli and Fang, Rongyao and Zhang, Yongfeng and Li, Hongsheng and Qiao, Yu},
  journal={International Journal of Computer Vision},
  volume={132},
  number={2},
  pages={581--595},
  year={2024}
}

@inproceedings{zhang2024cpl,
  title={Candidate Pseudolabel Learning: Enhancing Vision-Language Models by Prompt Tuning with Unlabeled Data},
  author={Zhang, Jiahan and Wei, Qi and Liu, Feng and Feng, Lei},
  booktitle={Proceedings of the 41st International Conference on Machine Learning},
  pages={60004--60020},
  year={2024}
}

@inproceedings{oh2024towards,
  title={Towards Calibrated Robust Fine-Tuning of Vision-Language Models},
  author={Oh, Changdae and Lim, Hyesu and Kim, Mijoo and Han, Dongyoon and Yun, Sangdoo and Choo, Jaegul and Hauptmann, Alexander and Cheng, Zhi-Qi and Song, Kyungwoo},
  booktitle={Advances in Neural Information Processing Systems},
  year={2024}
}

@article{tritraining,
  title={Tri-Training: Exploiting Unlabeled Data Using Three Classifiers},
  author={Zhou, Zhi-Hua and Li, Ming},
  journal={IEEE Transactions on Knowledge and Data Engineering},
  volume={17},
  number={11},
  pages={1529--1541},
  year={2005}
}

@article{learn-from-noisy-examples,
  title={Learning from Noisy Examples},
  author={Angluin, Dana and Laird, Philip},
  journal={Machine Learning},
  volume={2},
  number={4},
  pages={343--370},
  year={1988}
}

@inproceedings{wang2022debiased,
  title={Debiased Learning from Naturally Imbalanced Pseudo-Labels},
  author={Wang, Xudong and Wu, Zhirong and Lian, Long and Yu, Stella X},
  booktitle={Proceedings of the IEEE/CVF Conference on Computer Vision and Pattern Recognition},
  pages={14647--14657},
  year={2022}
}

@inproceedings{menghini2023grip,
  title={Enhancing {CLIP} with {CLIP}: Exploring Pseudolabeling for Limited-Label Prompt Tuning},
  author={Menghini, Cristina and Delworth, Andrew and Bach, Stephen},
  booktitle={Advances in Neural Information Processing Systems},
  year={2023}
}

@inproceedings{li2021comatch,
  title={Semi-Supervised Learning with Contrastive Graph Regularization},
  author={Li, Junnan and Xiong, Caiming and Hoi, Steven C. H.},
  booktitle={Proceedings of the IEEE/CVF International Conference on Computer Vision},
  year={2021}
}

@inproceedings{zhang2021flexmatch,
  title={{FlexMatch}: Boosting Semi-Supervised Learning with Curriculum Pseudo Labeling},
  author={Zhang, Bowen and Wang, Yidong and Hou, Wenxin and Wu, Hao and Wang, Jindong and Okumura, Manabu and Shinozaki, Takahiro},
  booktitle={Advances in Neural Information Processing Systems},
  year={2021}
}

@inproceedings{zenk2022realistic,
  title={Realistic Evaluation of {FixMatch} on Imbalanced Medical Image Classification Tasks},
  author={Zenk, Maximilian and Zimmerer, David and Isensee, Fabian and J{\"a}ger, Paul F and Wasserthal, Jakob and Maier-Hein, Klaus},
  booktitle={German Workshop on Medical Image Computing},
  pages={291--296},
  year={2022}
}

@inproceedings{codella2018skin,
  title={Skin Lesion Analysis Toward Melanoma Detection: A Challenge at the 2017 International Symposium on Biomedical Imaging ({ISBI}), Hosted by the International Skin Imaging Collaboration ({ISIC})},
  author={Codella, Noel CF and Gutman, David and Celebi, M Emre and Helba, Brian and Marchetti, Michael A and Dusza, Stephen W and Kalloo, Aadi and Liopyris, Konstantinos and Mishra, Nabin and Kittler, Harald and others},
  booktitle={Proceedings of the IEEE 15th International Symposium on Biomedical Imaging},
  year={2018}
}

@inproceedings{lee2013pseudolabel,
  title={Pseudo-Label: The Simple and Efficient Semi-Supervised Learning Method for Deep Neural Networks},
  author={Lee, Dong-Hyun and others},
  booktitle={Workshop on Challenges in Representation Learning, ICML},
  year={2013}
}

@inproceedings{zhang2022tipadapter,
  title={{Tip-Adapter}: Training-Free Adaption of {CLIP} for Few-Shot Classification},
  author={Zhang, Renrui and Zhang, Wei and Fang, Rongyao and Gao, Peng and Li, Kunchang and Dai, Jifeng and Qiao, Yu and Li, Hongsheng},
  booktitle={Proceedings of the European Conference on Computer Vision},
  pages={493--510},
  year={2022}
}

@article{zhou2022coop,
  title={Learning to Prompt for Vision-Language Models},
  author={Zhou, Kaiyang and Yang, Jingkang and Loy, Chen Change and Liu, Ziwei},
  journal={International Journal of Computer Vision},
  volume={130},
  number={9},
  pages={2337--2348},
  year={2022}
}

@inproceedings{zhou2022cocoop,
  title={Conditional Prompt Learning for Vision-Language Models},
  author={Zhou, Kaiyang and Yang, Jingkang and Loy, Chen Change and Liu, Ziwei},
  booktitle={Proceedings of the IEEE/CVF Conference on Computer Vision and Pattern Recognition},
  pages={16816--16825},
  year={2022}
}

@inproceedings{khattak2023maple,
  title={{MaPLe}: Multi-Modal Prompt Learning},
  author={Khattak, Muhammad Uzair and Rasheed, Hanoona and Maaz, Muhammad and Khan, Salman and Khan, Fahad Shahbaz},
  booktitle={Proceedings of the IEEE/CVF Conference on Computer Vision and Pattern Recognition},
  pages={19113-19122},
  year={2023}
}

@inproceedings{erasing-the-bias,
  title={Erasing the Bias: Fine-Tuning Foundation Models for Semi-Supervised Learning},
  author={Gan, Kai and Wei, Tong},
  booktitle={Proceedings of the 41st International Conference on Machine Learning},
  pages={14453--14470},
  year={2024}
}

@article{gupte2024vpet,
  title={Revisiting Active Learning in the Era of Vision Foundation Models},
  author={Gupte, Sanket Rajan and Aklilu, Josiah and Nirschl, Jeffrey J and Yeung-Levy, Serena},
  journal={Transactions on Machine Learning Research},
  year={2024}
}

@inproceedings{khattak2023promptsrc,
  title={Self-Regulating Prompts: Foundational Model Adaptation without Forgetting},
  author={Khattak, Muhammad Uzair and Wasim, Syed Talal and Naseer, Muzammal and Khan, Salman and Yang, Ming-Hsuan and Khan, Fahad Shahbaz},
  booktitle={Proceedings of the IEEE/CVF International Conference on Computer Vision},
  pages={15190--15200},
  year={2023}
}

@inproceedings{lv2025bmip,
  title={{BMIP}: Bi-Directional Modality Interaction Prompt Learning for {VLM}},
  author={Lv, Song-Lin and Chen, Yu-Yang and Zhou, Zhi and Yang, Ming and Guo, Lan-Zhe},
  booktitle={Proceedings of the International Joint Conference on Artificial Intelligence},
  pages={5887--5895},
  year={2025}
}

@article{unlabel_or_pretrain,
  title={Unlabeled Data vs. Pre-trained Knowledge: Rethinking SSL in the Era of Large Models},
  author={Lv, Song-Lin and Zhu, Rui and Wei, Tong and Li, Yu-Feng and Guo, Lan-Zhe},
  journal={arXiv preprint arXiv:2505.13317},
  year={2025}
}

@inproceedings{decoop,
  title={{DeCoOp}: Robust Prompt Tuning with Out-of-Distribution Detection},
  author={Zhou, Zhi and Yang, Ming and Shi, Jiang-Xin and Guo, Lan-Zhe and Li, Yu-Feng},
  booktitle={Proceedings of the 41st International Conference on Machine Learning},
  year={2024}
}

@inproceedings{rich_get_richer,
  title={The Rich Get Richer: Disparate Impact of Semi-Supervised Learning},
  author={Zhu, Zhaowei and Luo, Tianyi and Liu, Yang},
  booktitle={International Conference on Learning Representations},
  year={2022}
}

@inproceedings{debiased_self_training,
  title={Debiased Self-Training for Semi-Supervised Learning},
  author={Chen, Baixu and Jiang, Junguang and Wang, Ximei and Wan, Pengfei and Wang, Jianmin and Long, Mingsheng},
  booktitle={Advances in Neural Information Processing Systems},
  year={2022}
}

@inproceedings{vlm_peft_survey,
  title={Vision-Language Model Fine-Tuning via Simple Parameter-Efficient Modification},
  author={Li, Ming and Zhong, Jike and Li, Chenxin and Li, Liuzhuozheng and Lin, Nie and Sugiyama, Masashi},
  booktitle={Proceedings of the 2024 Conference on Empirical Methods in Natural Language Processing},
  pages={14394--14410},
  year={2024}
}

@article{yang2022sslsurvey,
  title={A Survey on Deep Semi-Supervised Learning},
  author={Yang, Xiangli and Song, Zixing and King, Irwin and Xu, Zenglin},
  journal={IEEE Transactions on Knowledge and Data Engineering},
  volume={35},
  number={9},
  pages={8934--8954},
  year={2023}
}

@article{huang2022upl,
  title={Unsupervised Prompt Learning for Vision-Language Models},
  author={Huang, Tony and Chu, Jack and Wei, Fangyun},
  journal={arXiv preprint arXiv:2204.03649},
  year={2022}
}

@inproceedings{fixmatch,
  title={{FixMatch}: Simplifying Semi-Supervised Learning with Consistency and Confidence},
  author={Sohn, Kihyuk and Berthelot, David and Carlini, Nicholas and Zhang, Zizhao and Zhang, Han and Raffel, Colin A and Cubuk, Ekin Dogus and Kurakin, Alexey and Li, Chun-Liang},
  booktitle={Advances in Neural Information Processing Systems},
  year={2020}
}

@article{cifar,
  title={Learning Multiple Layers of Features from Tiny Images},
  author={Krizhevsky, Alex and Hinton, Geoffrey and others},
  year={2009}
}

@inproceedings{shao2026chinatravel,
  title={{ChinaTravel}: An Open-Ended Travel Planning Benchmark with Compositional Constraint Validation for Language Agents},
  author={Shao, Jie-Jing and Zhang, Bo-Wen and Yang, Xiao-Wen and Chen, Baizhi and Han, Siyu and Pang, Jinghao and Wei, Wen-Da and Cai, Guohao and Dong, Zhenhua and Guo, Lan-Zhe and Li, Yu-Feng},
  booktitle={International Conference on Learning Representations},
  year={2026}
}

@inproceedings{yang2025neurosymbolic,
  title={Neuro-Symbolic Artificial Intelligence: Towards Improving the Reasoning Abilities of Large Language Models},
  author={Yang, Xiao-Wen and Shao, Jie-Jing and Guo, Lan-Zhe and Zhang, Bo-Wen and Zhou, Zhi and Jia, Lin-Han and Dai, Wang-Zhou and Li, Yu-Feng},
  booktitle={Proceedings of the International Joint Conference on Artificial Intelligence},
  pages={10770--10778},
  year={2025}
}

@article{li2021safeweakly,
  title={Towards Safe Weakly Supervised Learning},
  author={Li, Yu-Feng and Guo, Lan-Zhe and Zhou, Zhi-Hua},
  journal={IEEE Transactions on Pattern Analysis and Machine Intelligence},
  volume={43},
  number={1},
  pages={334--346},
  year={2021}
}

@inproceedings{guo2020safedeep,
  title={Safe Deep Semi-Supervised Learning for Unseen-Class Unlabeled Data},
  author={Guo, Lan-Zhe and Zhang, Zhen-Yu and Jiang, Yuan and Li, Yu-Feng and Zhou, Zhi-Hua},
  booktitle={Proceedings of the 37th International Conference on Machine Learning},
  pages={3897--3906},
  year={2020}
}

@article{guo2025robust,
  title={Robust Semi-Supervised Learning in Open Environments},
  author={Guo, Lan-Zhe and Jia, Lin-Han and Shao, Jie-Jing and others},
  journal={Frontiers of Computer Science},
  volume={19},
  pages={198345},
  year={2025}
}

@inproceedings{guo2022classimbalanced,
  title={Class-Imbalanced Semi-Supervised Learning with Adaptive Thresholding},
  author={Guo, Lan-Zhe and Li, Yu-Feng},
  booktitle={Proceedings of the 39th International Conference on Machine Learning},
  pages={8082--8094},
  year={2022}
}

@inproceedings{vmsr2025icml,
  title={Vision-Language Model Selection and Reuse for Downstream Adaptation},
  author={Tan, Hao-Zhe and Zhou, Zhi and Li, Yu-Feng and Guo, Lan-Zhe},
  booktitle={Proceedings of the 42nd International Conference on Machine Learning},
  pages={58726--58738},
  year={2025}
}

@inproceedings{imagenet,
  title={{ImageNet}: A Large-Scale Hierarchical Image Database},
  author={Deng, Jia and Dong, Wei and Socher, Richard and Li, Li-Jia and Li, Kai and Fei-Fei, Li},
  booktitle={Proceedings of the IEEE Conference on Computer Vision and Pattern Recognition},
  pages={248--255},
  year={2009}
}

@inproceedings{caltech101,
  title={Learning Generative Visual Models from Few Training Examples: An Incremental Bayesian Approach Tested on 101 Object Categories},
  author={Fei-Fei, Li and Fergus, Rob and Perona, Pietro},
  booktitle={Proceedings of the IEEE Conference on Computer Vision and Pattern Recognition Workshops},
  pages={178},
  year={2004}
}

@inproceedings{oxfordpets,
  title={Cats and Dogs},
  author={Parkhi, Omkar M and Vedaldi, Andrea and Zisserman, Andrew and Jawahar, CV},
  booktitle={Proceedings of the IEEE Conference on Computer Vision and Pattern Recognition},
  pages={3498--3505},
  year={2012}
}

@inproceedings{stanfordcars,
  title={{3D} Object Representations for Fine-Grained Categorization},
  author={Krause, Jonathan and Stark, Michael and Deng, Jia and Fei-Fei, Li},
  booktitle={Proceedings of the IEEE International Conference on Computer Vision Workshops},
  pages={554--561},
  year={2013}
}

@inproceedings{oxfordflowers,
  title={Automated Flower Classification over a Large Number of Classes},
  author={Nilsback, Maria-Elena and Zisserman, Andrew},
  booktitle={Proceedings of the 6th Indian Conference on Computer Vision},
  pages={722--729},
  year={2008}
}

@inproceedings{food101,
  title={{Food-101}: Mining Discriminative Components with Random Forests},
  author={Bossard, Lukas and Guillaumin, Matthieu and Van Gool, Luc},
  booktitle={Proceedings of the European Conference on Computer Vision},
  pages={446--461},
  year={2014}
}

@misc{fgvc,
  title={Fine-Grained Visual Classification of Aircraft},
  author={Maji, Subhransu and Rahtu, Esa and Kannala, Juho and Blaschko, Matthew and Vedaldi, Andrea},
  year={2013}
}

@inproceedings{sun397,
  title={{SUN} Database: Large-Scale Scene Recognition from Abbey to Zoo},
  author={Xiao, Jianxiong and Hays, James and Ehinger, Krista A and Oliva, Aude and Torralba, Antonio},
  booktitle={Proceedings of the IEEE Conference on Computer Vision and Pattern Recognition},
  pages={3485--3492},
  year={2010}
}

@misc{ucf101,
  title={{UCF}101: A Dataset of 101 Human Actions Classes from Videos in the Wild},
  author={Soomro, Khurram and Zamir, Amir Roshan and Shah, Mubarak},
  year={2012}
}

@inproceedings{dtd,
  title={Describing Textures in the Wild},
  author={Cimpoi, Mircea and Maji, Subhransu and Kokkinos, Iasonas and Mohamed, Sammy and Vedaldi, Andrea},
  booktitle={Proceedings of the IEEE Conference on Computer Vision and Pattern Recognition},
  pages={3606--3613},
  year={2014}
}

@article{eurosat,
  title={{EuroSAT}: A Novel Dataset and Deep Learning Benchmark for Land Use and Land Cover Classification},
  author={Helber, Patrick and Bischke, Benjamin and Dengel, Andreas and Borth, Damian},
  journal={IEEE Journal of Selected Topics in Applied Earth Observations and Remote Sensing},
  pages={2217--2226},
  year={2019}
}

\clearpage

\section{Supplementary Material}

\subsection{Detailed Theoretical Analysis}
\subsubsection{Proof of Lemma 1}
\begin{proof}
When the aggregated training data contains noisy samples, let $\mathcal{F}$ denote the hypothesis space, $\eta$ the noise ratio in the training set ($\eta < 0.5$), $\epsilon > 0$ the target error rate, and $\delta < 1$ the confidence parameter.
According to PAC learning theory \cite{learn-from-noisy-examples,tritraining}, if the number of training samples $m$ satisfies
\begin{equation}
m \geq \frac{2}{\epsilon^{2}(1-2 \eta)^{2}} \ln \left( \frac{2|\mathcal{F}|}{\delta} \right),
\label{eq:pac_noise}
\end{equation}
then, for the learned model $f$ and the optimal model $f^*$ within $\mathcal{F}$, we have
\begin{equation}
\mathbb{P}\left[ d(f, f^{*}) \geq \epsilon \right] \leq \delta,
\end{equation}
where $d(f, f^*)$ denotes the excess error rate of $f$ over $f^*$.

Let $c = 2\mu \ln \left( \frac{2|\mathcal{F}|}{\delta} \right)$, where $\mu$ makes is Eq. \eqref{eq:pac_noise} hold equality.
Eq. \eqref{eq:pac_noise} can then be rewritten as
\begin{equation}
m = \frac{c}{\epsilon^{2}(1 - 2\eta)^{2}}.
\end{equation}
Thus, the relationship between the error rate $\epsilon$, the number of training samples $m$, and the noise ratio $\eta$ is given by
\begin{equation}
\epsilon = \frac{\sqrt{c}}{\sqrt{m(1 - 2\eta)^{2}}}.
\label{eq:eps_relation}
\end{equation}

In the Bi-CoG training framework, we maintain $K$ vision-language models (VLMs). Suppose our optimization target is the $j$-th model, denoted as $f_j$. We adopt the $\textit{leave-one-out}$ scheme and generate a pseudo-label candidate set $D_{PL}$ using the remaining $K-1$ models via inter-model consistency and intra-model consistency. At iteration $t$, the candidate training set is
\begin{equation}
D_{train} = D_{L} \cup D_{PL},
\end{equation}
Let $e_t$ be the probability that the prediction obtained by a majority vote (agreed upon by more than half of the $K-1$ models) is still incorrect. Denote $|D_{L}| = L$ and $|D_{PL}| = L_t$, and approximate $|D_{\text{train}}| \approx L + L_t$. The noise ratio at iteration $t$ is then estimated as
\begin{equation}
\eta_t = \frac{e_t \times L_t}{L + L_t}.
\end{equation}

From Eq. \eqref{eq:eps_relation}, $\epsilon$ is inversely proportional to $\sqrt{m(1 - 2\eta)^2}$. Therefore, the condition $\epsilon_t < \epsilon_{t-1}$ is equivalent to

\begin{equation}
\begin{aligned}
(L + L_{t})&\left(1-2 \times \frac{e_{t}\times L_{t}}{L + L_{t}}\right)^{2} > \\
(L + L_{t-1})&\left(1-2 \times \frac{e_{t-1}\times L_{t-1}}{L + L_{t-1}}\right)^{2} \\
\end{aligned}
\end{equation}

A sufficient condition for $\epsilon_t < \epsilon_{t-1}$ is
\begin{equation}
\begin{cases}
L_t > L_{t-1}, \\
e_t \times L_t < e_{t-1} \times L_{t-1},
\end{cases}
\end{equation}
which can be equivalently expressed as
\begin{equation}
\label{important_equation}
0 < \frac{e_t}{e_{t-1}} < \frac{L_{t-1}}{L_t} < 1.
\end{equation}

Hence, we complete the proof.
\end{proof}

\subsubsection{Proof of Theorem 1}
\begin{proof}
Let $e_t$ denote the true relative error rate of pseudo-labels at iteration $t$, and let $\hat{e}_t$ denote its estimated value computed on the labeled dataset at the same iteration.
We begin by assuming that, during the course of training, the ratio of the true error rates between two consecutive iterations can be closely approximated by a power-law function of the corresponding ratio of the estimated error rates, given by:
\begin{equation}
\frac{e_{t}}{e_{t-1}} \approx \left(\frac{\hat{e}_{t}}{\hat{e}_{t-1}}\right)^{\alpha t}
\end{equation}
where $\alpha$ is a scaling factor characterizing the rate of error reduction.
Substituting this relationship into the condition in Eq. \eqref{important_equation}, we obtain
\begin{equation}
0 < \left(\frac{\hat{e}_{t}}{\hat{e}_{t-1}}\right)^{\alpha t} < \frac{L_{t-1}}{L_t} < 1
\end{equation}

\begin{table}[t]
\centering
    \fontsize{9}{12}\selectfont
    \setlength{\tabcolsep}{.5mm}
    \begin{tabular}{lcccc}
    \toprule
    Dataset & Domain & Categories & Train & Test  \\
    \midrule
     ImageNet & Natural picture & 1,000 & 1.28M & 50,000  \\
     Caltech101 & Object picture & 100 & 4,128 &  2,465 \\
     OxfordPets & Pet picture & 37 & 2,944 & 3,669  \\
     StanfordCars & Car picture & 196 & 6,509 &  8,041 \\
     Flowers102 & Flower picture & 102 & 4,093 & 2,463  \\
     Food101 & Food picture & 101 & 50,500 & 30,300  \\
     FGVCAircraft & Aircraft picture & 100 & 3,334 & 3,333  \\
     SUN397 & Scene picture & 397 & 15,880 & 19,850  \\
     DTD & Texture picture & 47 & 2,820 & 1,692  \\
     EuroSAT & Satellite picture & 10 & 13,500 &  8,100 \\
     UCF101 & Video frame & 101 & 7,639 & 3,783  \\
     CIFAR-10 & Natural picture & 10  & 50,000 & 10,000  \\
     CIFAR-100 & Natural picture & 100 & 50,000 & 10,000  \\
     ImageNet-100 & Natural picture & 100 & 130,000 & 5,000  \\
    \bottomrule
    \end{tabular}
    \caption{Details on the datasets used in our experiments.}
    \label{tab:dataset_details}
\end{table}

Therefore, the maximum allowable number of pseudo-labeled samples at iteration $t$ is given by
\begin{equation}
L_{t}^* = \left\lceil \left(\frac{\hat{e}_{t-1}}{\hat{e}_{t}}\right)^{\alpha t} \times L_{t-1} - 1 \right\rceil
\end{equation}

In addition, the number of pseudo-labeled samples at iteration $t-1$ must satisfy
\begin{equation}
L_{t-1} < \left\lceil \left(\frac{\hat{e}_{t-1}}{\hat{e}_{t}}\right)^{\alpha t} \times L_{t-1} - 1 \right\rceil
\end{equation}
which is equivalent to
\begin{equation}
L_{t-1} > \frac{\hat{e}_{t}^{\alpha t}}{\hat{e}_{t-1}^{\alpha t} - \hat{e}_{t}^{\alpha t}}
\end{equation}

Consequently, we obtain both the upper bound on the number of pseudo-labeled samples at iteration $t$ and the corresponding lower bound constraint on $L_{t-1}$.
\end{proof}

\subsection{Implementation Details}

\begin{table*}[t]
\centering
    \fontsize{10}{12}\selectfont
    \setlength{\tabcolsep}{.8mm}
\begin{tabular}{lccc}
\toprule
    \textbf{Transformation} & \textbf{Description} & \textbf{Parameter} & \textbf{Value} \\ \midrule
\rowcolor{gray!20}
\multicolumn{4}{l}{\textit{Weak Augmentations}} \\ \midrule
\texttt{RandomHorizontalFlip} & Flips the image horizontally with a probability of $p$. & $p$ & $0.5$ \\ \midrule
\multirow{2}{*}{\texttt{RandomCrop}} & \multirow{2}{*}{Crops a random patch of the image after padding.} & size & (224, 224) \\
 & & padding & 28 \\ \midrule
 \rowcolor{gray!20}
\multicolumn{4}{l}{\textit{Strong Augmentations}} \\ \midrule
\multirow{4}{*}{\texttt{ColorJitter}} & \multirow{4}{*}{Randomly changes the brightness, contrast, saturation, and hue.} & brightness & 0.4\\
 & & contrast & 0.4\\
 & & saturation & 0.1\\
 & & hue & 0.4\\ \midrule
\texttt{RandomGrayscale} & Converts the image to grayscale with a probability of $p$. & $p$ & $0.2$ \\ \midrule
\multirow{2}{*}{\texttt{GaussianBlur}} & \multirow{2}{*}{Applies Gaussian blur to the image with a probability of $p$.} & kernel\_size & 21 \\
 & & $p$ & 0.5 \\ \bottomrule
\end{tabular}
\caption{Details of Weak and Strong Data Augmentations.}
\label{tab:augmentations}
\end{table*}

\paragraph{Dataset Details.} We evaluate our method on 14 recognition datasets, comprising 11 commonly used benchmarks for evaluating vision-language models (VLMs) and 3 classical benchmarks for semi-supervised learning.
These datasets include CIFAR-10 and CIFAR-100~\cite{cifar}, ImageNet and ImageNet-100~\cite{imagenet}, Caltech101~\cite{caltech101}, OxfordPets~\cite{oxfordpets}, StanfordCars~\cite{stanfordcars}, Food101~\cite{food101}, Flowers102~\cite{oxfordflowers}, FGVCAircraft~\cite{fgvc}, SUN397~\cite{sun397}, UCF101~\cite{ucf101}, DTD~\cite{dtd}, and EuroSAT~\cite{eurosat}.

For ImageNet, Caltech101, OxfordPets, StanfordCars, Flowers102, FGVCAircraft, SUN397, UCF101, DTD, and EuroSAT, we follow the dataset splits used in CoOp~\cite{zhou2022cocoop}. For CIFAR-10, CIFAR-100, and ImageNet-100, we adopt the standard semi-supervised splits used in FixMatch~\cite{fixmatch}. Detailed statistics of these datasets are provided in Table~\ref{tab:hyperparameters}. Specifically, for the open-world generalization experiment, we follow the standard base-to-novel setting\cite{zhou2022cocoop}, where base and novel classes are split evenly. We sample 16 shots per class as labeled data, and treat all remaining training samples as unlabeled. Due to computational constraints, for ImageNet we randomly sample 5\% of the unlabeled set for training.

\paragraph{Training Details.}

As a plug-and-play method, Bi-CoG can be seamlessly integrated into any pretrain--finetune paradigm.
In our experiments, we adopt CoOp~\cite{zhou2022coop}, MaPLe~\cite{khattak2023maple}, and PromptSRC~\cite{khattak2023promptsrc} as the base methods, keeping all hyperparameters identical to those reported in their original papers.
Unless otherwise specified, we set the number of models to $K=3$ and the scaling factor to $\alpha=1$.
In the dynamic self-training stage, each model is first warmed up for $\zeta$ epochs using a small set of labeled data.
Then, we apply our three-stage pseudo-label selection strategy to obtain pseudo-labels with high accuracy and low model bias, which are further used for self-training over $\kappa$ additional epochs.
The detailed parameters are provided in Table \ref{tab:hyperparameters}.
During the intra-model consistency stage, we apply weak and strong augmentations to each sample, with the specific augmentation strategies summarized in Table \ref{tab:augmentations}. We conduct all experiments on a single NVIDIA A800 GPU.

\begin{table}[t]
\centering
    \fontsize{9}{12}\selectfont
    \setlength{\tabcolsep}{.8mm}
\begin{tabular}{lcc|cc|cc}
\toprule
\multirow{2}{*}{Dataset} & \multicolumn{2}{c|}{CoOp} & \multicolumn{2}{c|}{MaPLe} & \multicolumn{2}{c}{PromptSRC} \\ \cline{2-7}
                         & $\zeta$       & $\kappa$      & $\zeta$       & $\kappa$      & $\zeta$         & $\kappa$        \\ \midrule
CIFAR-100                & 15            & 1             & 5             & 1             & 15              & 1               \\
ImageNet                 & 15            & 3             & 5             & 1             & 15              & 1               \\
FGVCAircraft             & 15            & 3             & 5             & 1             & 20              & 1               \\
SUN397                   & 15            & 3             & 5             & 1             & 15              & 1               \\
UCF101                   & 15            & 3             & 5             & 1             & 15              & 1               \\
CIFAR-10                 & 15            & 3             & 5             & 1             & 15              & 3               \\
DTD                      & 15            & 3             & 5             & 1             & 15              & 1               \\
EuroSAT                  & 15            & 3             & 7             & 1             & 5               & 20              \\
StanfordCars             & 15            & 3             & 7             & 1             & 5               & 20              \\
ImageNet-100             & 15            & 5             & 5             & 1             & 15              & 3               \\
Caltech101               & 20            & 5             & 5             & 1             & 20              & 5               \\
OxfordPets               & 20            & 5             & 7             & 1             & 15              & 1               \\
Flowers102               & 20            & 5             & 5             & 1             & 20              & 5               \\
Food101                  & 20            & 5             & 5             & 1             & 15              & 1               \\ \bottomrule
\end{tabular}
\caption{Details of the training epoch.}
\label{tab:hyperparameters}
\end{table}

\subsection{Additional Experiments}
\begin{table}[t]
    \centering

    \begin{tabular}{lc}
    \toprule
    \textbf{Method} & \textbf{Accuracy (\%)} \\
    \midrule
    Self-training & 95.02 \\
    FlexMatch (RA) & 93.40 $\pm$ 0.05 \\
    CoMatch (RA) & 93.39 $\pm$ 0.04 \\
    FixMatch (RA) & 93.23 $\pm$ 0.14 \\
    \midrule
    Zero-shot CLIP & 14.60 \\
    PromptSRC & 59.97 $\pm$ 2.09 \\
    PromptSRC + Bi-CoG & 61.03 $\pm$ 2.03 \\
    \bottomrule
    \end{tabular}
    \caption{Performance comparison of different methods on the ISIC2018 medical image dataset.}
    \label{tab:isic2018}
\end{table}

To further validate the effectiveness of Bi-CoG in domain-specific scenarios with limited labeled data, we conduct experiments on the ISIC2018 dataset \cite{codella2018skin}, which focuses on skin lesion diagnosis and comprises annotated images across seven distinct categories of skin diseases.

For conventional SSL methods (Self-training, FlexMatch \cite{zhang2021flexmatch}, CoMatch \cite{li2021comatch}, and FixMatch \cite{fixmatch}), we directly report the results from \cite{zenk2022realistic} to ensure consistency with established experimental benchmarks. For Bi-CoG, we adopt PromptSRC \cite{khattak2023promptsrc} as the base model---an exemplar prompt tuning method tailored for vision-language models (VLMs). For statistical robustness, all results for Bi-CoG and PromptSRC are averaged over three runs with distinct random seeds.
As illustrated in Table~\ref{tab:isic2018}, Zero-shot CLIP achieves a mere 14.60\% accuracy on this dataset. This poor performance stems from the paucity of medical image samples in CLIP's pre-training corpus, leading to a severe mismatch between its generic visual features and the specialized requirements of skin lesion diagnosis. After prompt-based fine-tuning, PromptSRC significantly boosts the accuracy to 59.97\%, demonstrating the effectiveness of task-specific prompt tuning in adapting VLMs to medical domains. When integrated with Bi-CoG, PromptSRC further yields 1.06\% improvement to 61.03\%, verifying that Bi-CoG's adaptive pseudo-labeling strategy can effectively refine model performance even in specialized medical scenarios.

Notably, conventional SSL methods exhibit a substantial performance lead over fine-tuned VLMs on this dataset, with Self-training achieving the highest accuracy of 95.02\%. This stark performance gap highlights a critical insight: constrained by the inherent limitations of their generic pre-trained features, VLMs still lag far behind end-to-end SSL algorithms in data-scarce medical imaging domains. Thus, for scenarios where labeled data is extremely limited and domain specificity is strong, it is more advisable to adopt traditional end-to-end SSL approaches rather than relying solely on pre-trained VLMs for downstream adaptation.

\end{document}